\documentclass[10pt,twocolumn,letterpaper]{article}

\usepackage[algorithms]{wacv}      % To produce the REVIEW version for the algorithms track

\usepackage{times}
\usepackage{epsfig}
\usepackage{graphicx}
\usepackage{amsmath}
\usepackage{amssymb}
\usepackage{booktabs}
\usepackage{multirow}

\usepackage{color, colortbl}
\usepackage{xcolor}
\usepackage{tikz}
\definecolor{orange}{rgb}{0.99,0.3,0.01}

\definecolor{gold}{HTML}{BD820B}%{EEAD0E}
\definecolor{silver}{HTML}{909090}%{C0C0C0}
\definecolor{bronze}{HTML}{9A5F26}%{CD7F32}

\newcommand*\circledd[1]{\tikz[baseline=(char.base)]{
            \node[shape=circle,draw,inner sep=0.15pt] (char) {#1};}}      

\newcommand{\first}[1]{%
    {#1\raisebox{0.8pt}{\footnotesize \color{gold} \circledd{1}}}%
}
\newcommand{\second}[1]{%
    {#1\raisebox{0.8pt}{\footnotesize \color{silver} \circledd{2}}}%
}
\newcommand{\third}[1]{%
    {#1\raisebox{0.8pt}{\footnotesize \color{bronze} \circledd{3}}}%
}

\usepackage[pagebackref,breaklinks,colorlinks]{hyperref}

\usepackage[capitalize]{cleveref}
\crefname{section}{Sec.}{Secs.}
\Crefname{section}{Section}{Sections}
\Crefname{table}{Table}{Tables}
\crefname{table}{Tab.}{Tabs.}

\def\wacvPaperID{***} % *** Enter the WACV Paper ID here
\def\confName{WACV}
\def\confYear{2024}

\begin{document}

%%%%%%%%% TITLE
%\title{Cheating depth -- Depth-aware Discrete Autoencoder for 3D anomaly detection}
\title{Cheating Depth: Enhancing 3D Surface Anomaly Detection via Depth Simulation}

\author{Vitjan Zavrtanik \and Matej Kristan \and Danijel Skočaj  \and
Faculty of Computer and Information Science, University of Ljubljana\\
%Večna pot 113\\
{\tt\small \{vitjan.zavrtanik, matej.kristan, danijel.skocaj\}@fri.uni-lj.si}
%\and
}

\maketitle
%\thispagestyle{empty}

%%%%%%%%% ABSTRACT
\begin{abstract}
RGB-based surface anomaly detection methods have advanced significantly.
However, certain surface anomalies remain practically invisible in RGB alone, necessitating the incorporation of 3D information. Existing approaches that employ point-cloud backbones suffer from suboptimal representations and reduced applicability due to slow processing. Re-training RGB backbones, designed for faster dense input processing, on industrial depth datasets is hindered by the limited availability of sufficiently large datasets. 
We make several contributions to address these challenges. (i)  
We propose a novel Depth-Aware Discrete Autoencoder (DADA) architecture, that enables learning a general discrete latent space that jointly models RGB and 3D data for 3D surface anomaly detection. 
(ii) We tackle the lack of diverse industrial depth datasets by introducing a simulation process for learning informative depth features in the depth encoder. 
(iii) We propose a new surface anomaly detection method 3DSR, which outperforms all existing state-of-the-art on the challenging MVTec3D anomaly detection benchmark, both in terms of accuracy and processing speed. The experimental results validate the effectiveness and efficiency of our approach, highlighting the potential of utilizing depth information for improved surface anomaly detection. Code is available at: \href{https://github.com/VitjanZ/3DSR}{https://github.com/VitjanZ/3DSR}
   %Top performing anomaly detection methods commonly rely on large pretrained neural networks that are able to extract informative features from RGB images. Such methods perform poorly on industrial 3D data from which pretrained backbones can not extract meaningful features. Some discriminative anomaly detection methods do not require a large pretrained network and are instead trained from scratch on anomaly-free data relying on simulated anomalies for supervision. Anomaly simulation can be done by sampling from a discrete feature space of a small quantized autoencoder. These quantized autoencoders do not produce semantically meaningful features but instead learn to represent local patch information. We hypothesize that learning a general industrial depth representation with quantized autoencoders does not require a large dataset for training and can in fact be trained with a simulated 3D dataset.  We propose a self-supervised pretraining process and a quantized autoencoder architecture to learn a general industrial depth representation enabling accurate downstream 3D anomaly detection. The proposed method achieves state-of-the-art results on the MVTec 3D anomaly detection dataset and the Eyecandies anomaly detection dataset, significantly outperforming the previous state-of-the-art in 3D and RGB+3D settings.
\end{abstract}

%%%%%%%%% BODY TEXT
\section{Introduction}

\begin{figure}[htbp]
  \centering
  \includegraphics[width=1.0\linewidth]{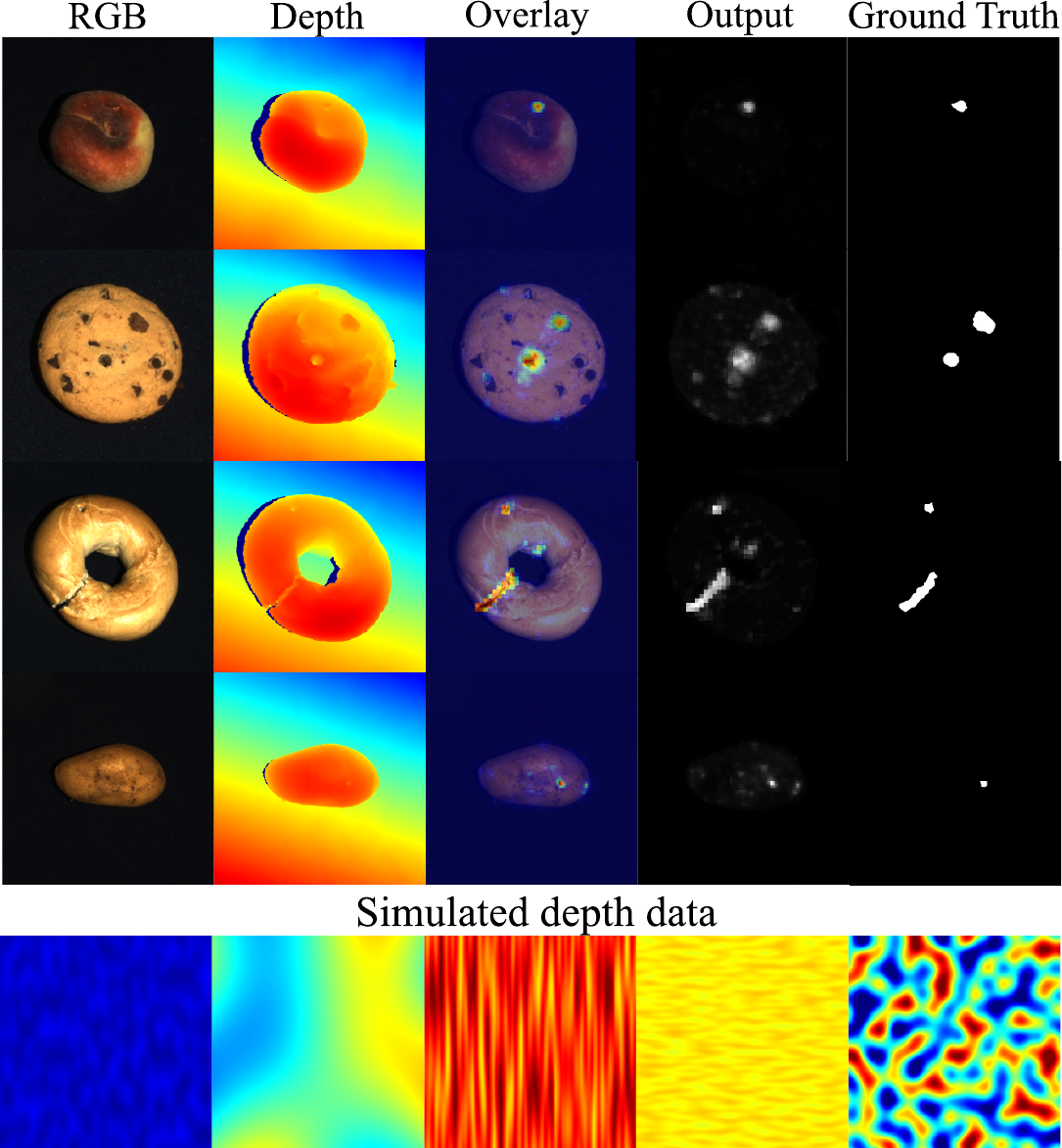}
  \caption{Certain anomalies are practically imperceptible in RGB, requiring depth for precise detection. Parameterized generative model yields images sufficiently describing depth statistics for training general depth-reconstruction backbones.}
  \label{fig:first}
\end{figure}

Surface anomaly detection addresses localization of image regions that deviate from normal object appearance.
Most of the works consider the problem of RGB-based detection, which
has witnessed remarkable progress in recent years, with several methods approaching perfection on the widely adopted MVTec anomaly detection dataset ~\cite{bergmann2019mvtec}. 
However, since certain surface anomalies in practical applications are not detectable in RGB (Figure~\ref{fig:first}), recent works~\cite{mvtec3d,rudolph2023ast,bergmann2023ST,horwitz2022empirical} consider a new research problem of RGB+3D anomaly detection. 

The state-of-the-art methods typically rely on features extracted by general backbone  networks~\cite{roth2021towards,rudolph2021same,rudolph2023ast,yang2023memseg}, pretrained on large datasets. The current state-of-the-art 3D anomaly detection method M3DM~\cite{wang2023m3dm} applies two backbones, one pretrained on RGB, the other on point-cloud datasets. However, the general point-cloud datasets do not represent well the depth appearance distribution of industrial setups, leading to suboptimal representations. Furthermore, the point-cloud backbone substantially slows down processing, reducing the method's practicality. The inference could be sped up by considering depth image as a grayscale image and replacing the point-cloud backbone with an RGB-pretrained one, but evidence shows~\cite{horwitz2022empirical} that RGB-pretrained backbones insufficiently represent the depth properties relevant for anomaly detection. 

Alternatively, RGB backbones could be re-trained on industrial depth datasets, but the current industrial depth datasets are too small to efficiently train the large backbones. 
Recent work DSR~\cite{zavrtanik2022dsr} proposed utilizing Vector-quantized autoencoders (VQVAE)~\cite{vqvae2}, which learn only a fixed number of discrete latent representation vectors, thus potentially enable learning from smaller datasets. 
Nevertheless, our experience shows that training DSR on the available industrial depth dataset MVTec3D~\cite{mvtec3d} leads to suboptimal results, indicating that the existing data is too small even for the representation-efficient VQVAEs. Advances in RGB+3D surface anomaly detection are thus hindered by the lack of sufficiently large datasets that would enable pre-training general depth backbones and allow development of methods fast enough for practical applications.

We address the aforementioned issues by making an insight that it is possible to summarize statistical properties of spatial and intensity content typical for industrial surface anomaly inspection depth data, which allows the creation of a parameterized generative models for such data. We hypothesize that such a model can then be used to generate a large training set for pre-training a depth-specific backbone (Figure~\ref{fig:first}, bottom). %However, care has to be taken 
Furthermore, we note that the processing time constraint requires efficient architecture design for encoding of the RGB and depth data to jointly exploit both modalities for detecting complex anomalies. %, allowing very fast processing times.

The main contributions of this work are three-fold: (i) A novel Depth-Aware Discrete Autoencoder (\textit{DADA}) architecture is proposed, that enables learning a general discrete latent space that jointly models RGB and depth data for 3D surface anomaly detection. (ii) We directly address the lack of a diverse industrial depth dataset by proposing an industrial depth data simulation process that facilitates the learning of informative depth features by the DADA module. (iii) We demonstrate the effectiveness of the learned feature space by proposing \textit{3DSR}, a novel discriminative 3D surface anomaly detection method that significantly outperforms all competing state-of-the-art methods on the most challenging MVTec3D anomaly detection benchmark~\cite{mvtec3d}. Owing to its efficient design, 3DSR also outperforms the current state-of-the-art~\cite{wang2023m3dm} in speed by by an order of magnitude.

\section{Related work}

The MVTec anomaly detection dataset~\cite{bergmann2019mvtec} has been widely adopted for unsupervised surface anomaly detection research. The training dataset comprises only normal cases, while the testing dataset includes both normal and anomalous instances. Most top performing anomaly detection methods~\cite{roth2021towards, rudolph2021same,defard2021padim,rudolph2023ast,yang2023memseg} rely on strong pretrained backbones for informative feature extraction. After obtaining a good representation of each anomaly-free image in the training set a simple statistical model is typically built that tightly binds the anomaly-free feature space~\cite{roth2021towards,defard2021padim}. This enables the detection of anomalies based on a chosen distance function to the anomaly-free representation model. Flow-based methods that utilize pretrained feature extractors to train a flow model also achieve state-of-the-art results~\cite{rudolph2022csflow,rudolph2021same,rudolph2023ast,yu2021fastflow}. Certain discriminative methods~\cite{zavrtanik2021draem, li2021cutpaste,zavrtanik2022dsr} do not need a strong pretrained backbone but instead rely on simulated anomalies using an out-of-distribution dataset to build a strong classifier that generalizes well to real anomalies at test-time. It has been shown that top-performing RGB anomaly detection methods do not generalize well to 3D surface anomaly detection~\cite{mvtec3d,rudolph2023ast,horwitz2022empirical}.

In the problem setup of 3D surface anomaly detection, the MVTec-3D anomaly detection dataset~\cite{mvtec3d} is the most comprehensive dataset containing RGB and 3D information. In \cite{mvtec3d} reconstruction-based anomaly detection methods have been applied to the 3D anomaly detection problem as an initial baseline. In \cite{bergmann2023ST}, a 3D student-teacher network was proposed that focuses on point cloud geometry descriptors. In \cite{horwitz2022empirical} a memory-based method akin to \cite{roth2021towards} was proposed, that utilizes pretrained backbone features for RGB and FPFH~\cite{rusu2009fast} features for 3D representation. In \cite{rudolph2023ast} a teacher-student flow-based model is proposed that uses pretrained backbone features for RGB data but raw depth pixel values are used for 3D representation. In \cite{wang2023m3dm} 3D features are extracted by a point transformer~\cite{zhao2021point} pretrained on a general point cloud dataset not adapted to industrial 3D data. A lack of 3D data from the industrial domain and the lack of strong domain-specific feature extractors presents a significant challenge since better 3D representations are required for improving the accuracy of 3D surface anomaly detection methods. % Knowledge distillation [27] further enhanced the detection of pure RGB and multimodal anomalies by incorporating depth information. In contrast to previous approaches, our method builds upon memory banks and introduces a novel pipeline that utilizes a pretrained point transformer and a hybrid feature fusion scheme to achieve more precise detection. We have extended the memory bank method to encompass 3D and multimodal settings, yielding impressive results.

\section{Our approach: 3DSR}

We propose a novel 3D surface anomaly detection method based on dual subspace reprojection (3DSR). The input image is encoded into a discrete feature space and is then reprojected into image space by two decoders. The object-specfic decoder and the general object decoder reconstruct the anomaly-free and the anomalous appearance, respectively. The anomaly detection module then segments potential anomalies based on the difference between the two reprojections. 3DSR is trained in two stages. First, a novel Depth-Aware Discrete Autoencoder (DADA) is trained on RGB and depth image pairs to learn a general joint discrete representation of RGB+3D depth data. 

In the second stage, DADA is integrated into the DSR~\cite{zavrtanik2022dsr} surface anomaly detection framework, producing 3DSR, which is then trained on 3D anomaly detection datasets~\cite{mvtec3d,bonfiglioli2022eyecandies}. In Section~\ref{sec:vqvae} the architecture of the proposed Depth-Aware Discrete Autoencoder (DADA) module is described. The industrial depth data simulation process is then described in Section~\ref{sec:depthgen}. The final 3DSR surface anomaly detection pipeline is described in Section~\ref{sec:3dsr}.

\subsection{Depth-aware discrete Autoencoder}\label{sec:vqvae}
\begin{figure}[htbp]
  \centering
  \includegraphics[width=1.0\linewidth]{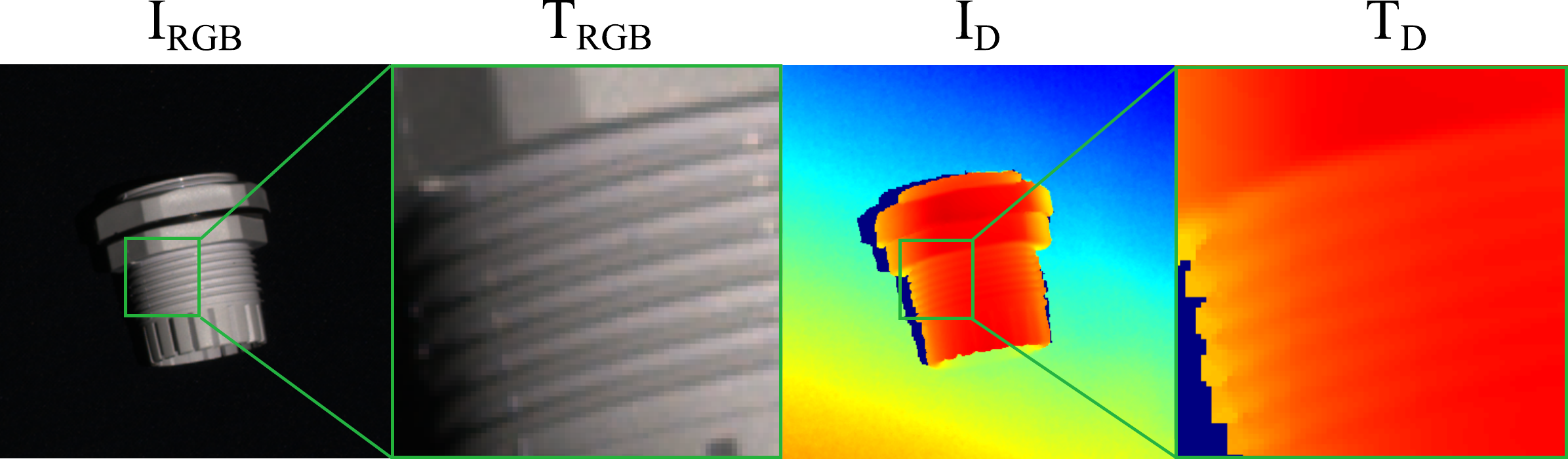}
  \caption{Example of a cable gland from the MVTec3D dataset~\cite{mvtec3d}. Local variations are clearly visible in $T_{RGB}$ and barely perceptible in $T_D$.}
  \label{fig:depth_ex}
\end{figure}

To learn a representation of both RGB and depth data, a naive approach may be to train a vector quantized autoencoder~\cite{vqvae2} with 4 input channels to represent both RGB and depth. This approach has some drawbacks due to the properties of both RGB and depth images. In Figure~\ref{fig:depth_ex} an example of a cable gland is shown. A region of interest is marked with a green rectangle in $I_{RGB}$ and $I_D$ and is shown in more detail in RGB ($T_{RGB}$) and depth ($T_{D}$). $T_{RGB}$ exhibits significant local variation due to shadows but barely any variation is visible in $T_{D}$. In depth images, even slight depth variations can be informative for defect detection so representing variations in $T_{D}$ is vital. Discrete autoencoders are typically trained using the $L2$ loss, which is less sensitive to variation types in $T_{D}$, where values change minimally, but is sensitive to changes in $T_{RGB}$, where local gradients are higher. Subtle changes in $T_{D}$ therefore contribute very little to the final loss leading to a higher emphasis on the reconstruction of RGB data during training. 

\begin{figure}[htbp]
  \centering
  \includegraphics[width=0.8\linewidth]{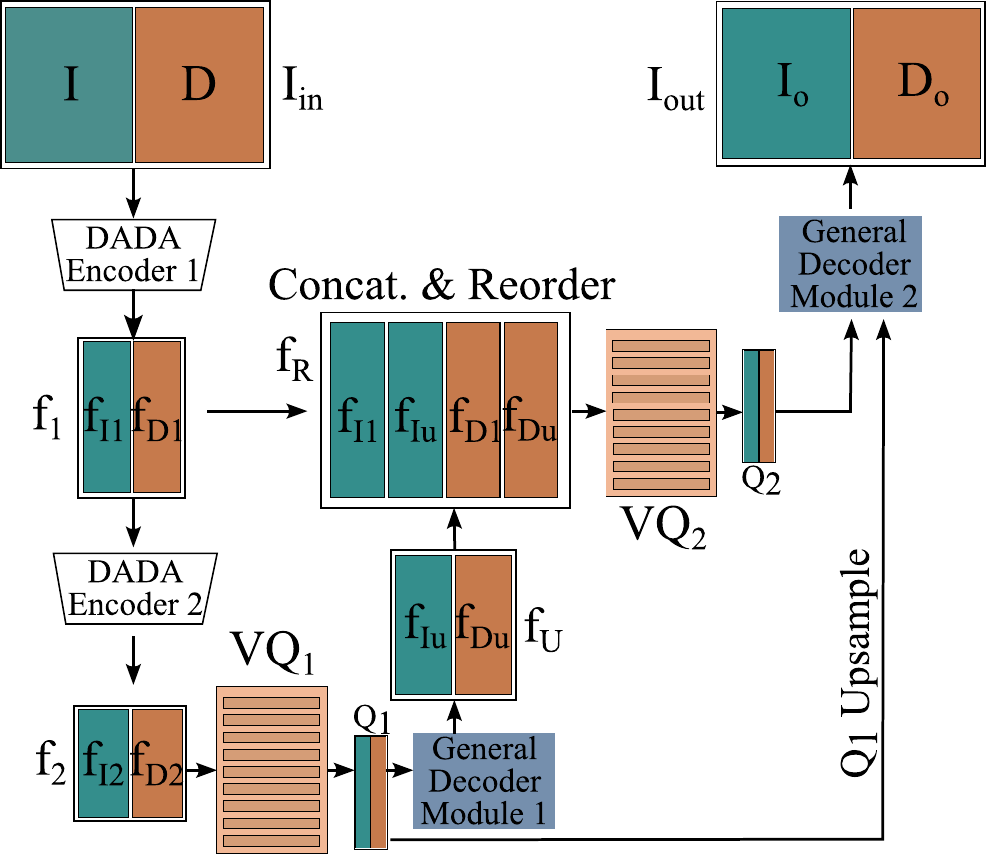}
  \caption{The Depth-Aware Discrete Autoencoder (DADA) module.}
  \label{fig:vqvae}
\end{figure}

To ensure a good representation of both RGB and depth, a new Depth-Aware Discrete Autoencoder (DADA) architecture is required and is shown in Figure~\ref{fig:vqvae}. The input $I_{in}$ to the discrete autoencoder is a 4 channel tensor, a concatenation of an RGB image $I$ and a depth image $D$. The encoder is a convolutional network, where RGB and depth information is separated by grouped convolution layers~\cite{hinton2012imagenet} which ensure that RGB and Depth features do not interact. This channel-wise separation is necessary to prevent the overwhelming influence of a single modality in the loss function. 

To minimize the information loss due to discretization at a low spatial resolution, a two-stage discretization architecture is used~\cite{vqvae2}. First, the DADA Encoder 1 encodes the input and downsamples the spatial resolution by $4\times$ producing features $f_1$, where $f_{I1}$ and $f_{D1}$ stand for RGB and depth features, respectively. The second encoder stage, DADA Encoder 2, further downsamples the features to a total $8\times$ downsampling, producing $f_{2}$. $f_{2}$ features are quantized to the nearest neighbors of the codebook $VQ_1$ in terms of the L2 distance. The quantized features $Q_1$ are then input into a decoder module which upsamples the spatial resolution $2\times$ to $f_{U}$. The features $f_1$ and $f_U$ are then concatenated and channel-wise reordered to group image features $f_{I1}$ and $f_{IU}$ and depth features $f_{D1}$ and $f_{Du}$. This reordering of $f_R$ is necessary to maintain the separation of RGB and depth features in the grouped convolutions of the decoder. The resulting features $f_R$ are then quantized to nearest neighbor codebook vectors in $VQ_2$, producing $Q_2$. $Q_1$ is then upsampled to fit the spatial resolution of $Q_2$, after which $Q_1$ and $Q_2$ are input into the second decoder module which outputs the reconstructions of RGB $I_o$ and depth $D_o$ concatenated as $I_{out}$. We use the VQ-VAE~\cite{vqvae2} loss function, modified to address the added depth information, to train DADA:
\begin{multline}
    \mathcal{L}_\mathrm{ae} = \lambda_{D} L_2(\mathbf{D}, \mathbf{D}_\mathrm{o}) + \lambda_{I} L_2(\mathbf{I}, \mathbf{I}_\mathrm{o}) \\
    + L_2(sg[\mathbf{f_2}], \mathbf{Q_1}) + \lambda_K L_2(\mathbf{f_2}, sg[\mathbf{Q_1}]) \\
    + L_2(sg[\mathbf{f_1}], \mathbf{Q_2}) + \lambda_K L_2(\mathbf{f_1}, sg[\mathbf{Q_2}]),
    \label{eq:vqv}
\end{multline}
where $L_2(\cdot)$ is the Euclidean distance and $sg[\cdot]$ is the stop gradient operator. $\lambda_{I}$ and $\lambda_{D}$ are loss weighing factors and are both set to $1$ in all experiments unless stated otherwise. $\lambda_K$ controls the reluctance to change the codebook vectors corresponding to $f_1$ and is fixed to $0.25$ in all experiments following \cite{vqvae2}. 

\subsection{Depth data generative model} % Industrial depth data simulation
\label{sec:depthgen}
Simulated data is necessary for training DADA due to the absence of industrial depth datasets. Consideration of key properties of industrial depth data is required for an effective simulation process. Firstly, object depth can vary continuously from closest to farthest from the sensor. Secondly, small dents and bumps can either cause significant intensity changes in RGB or are completely invisible, depending on the lighting. In depth, such minor changes are always detectable through a minimal local depth value alteration. Finally, the average of a depth image can vary significantly. Simulated data must capture local changes and variable average object depth in industrial images. The simulated depth image generation process is thus designed to explicitly address these properties. The core generator of the simulated images is the Perlin noise generator~\cite{perlin1985image} that produces a variety of locally smooth textures that simulate the gradual changes in depth well, addressing the first property. Subtle local changes and varied average object distance are then simulated by adapting the Perlin noise image with a randomized affine transform.

To generate a single simulated depth image a Perlin noise image $P$ is first generated and normalized between $0$ and $1$. $P$ is then multiplied by a uniformly sampled $\alpha \in (0.0,1.0)$ to produce $P_{\alpha}$. At lower $\alpha$ values the maximum and minimum values of $P_{\alpha}$ will differ only slightly. To model the variation of the average object distance, $P_{\alpha}$ is translated with a uniformly sampled $\beta \in (0, 1-\alpha)$. The final simulated depth image $D$ is therefore $D = \alpha P + \beta$. 
 
\begin{figure}[htbp]
  \centering
  \includegraphics[width=0.8\linewidth]{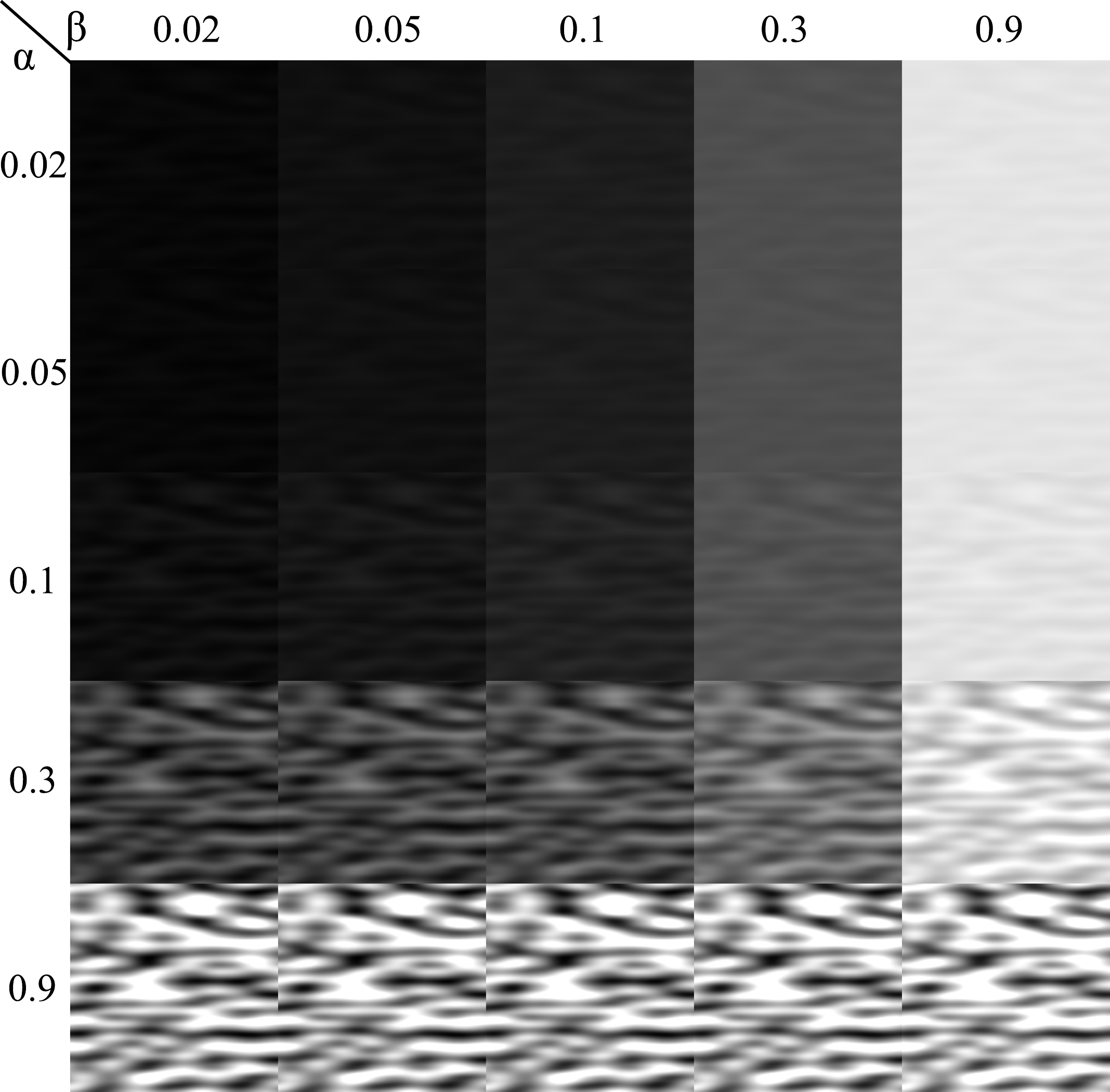}
  \caption{Impact of parameters $\alpha$ and $\beta$ on the simulated depth maps. $\alpha$ is the scaling parameter and $\beta$ is the translation parameter.}
  \label{fig:perlin}
\end{figure}

 Examples of simulated depth images with various $\alpha$ and $\beta$ parameters are shown in Figure~\ref{fig:perlin}. $\alpha$ controls the difference between the minimum and maximum values of $D$, simulating local changes in depth and $\beta$ controls the minimum value of $D$.

\begin{figure*}[htbp]
  \centering
  \includegraphics[width=1.0\linewidth]{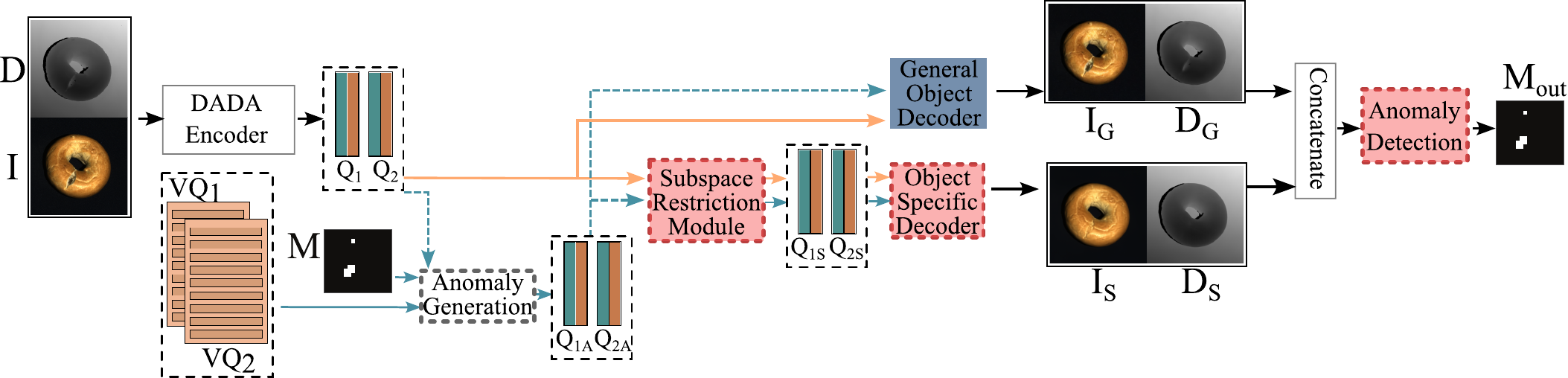}
  \caption{The architecture and second stage training process of 3DSR. The modules marked in red are trainable during the second stage of training.}
  \label{fig:3dsr}
\end{figure*}

DADA is then trained on RGB and depth image pairs. RGB images are sampled from ImageNet\cite{deng2009imagenet} and depth images are simulated (Figure \ref{fig:perlin}) and unrelated to the input RGB data.

\subsection{3D anomaly detection pipeline} \label{sec:3dsr}
In the second stage, DSR~\cite{zavrtanik2022dsr} is used as a discriminative anomaly detection framework. The VQ-VAE-2~\cite{vqvae2} network that is used by DSR for RGB surface anomaly detection is replaced with DADA, pretrained to extract informative representations from both 3D and RGB data. Additionally, DADA's vector-quantized feature space enables efficient simulated anomaly sampling. The architecture of 3DSR is shown in Figure~\ref{fig:3dsr}. 

The DADA encoder modules extract and quantize the features $Q_1$ and $Q_2$. Then $Q_1$ and $Q_2$ are modified by the anomaly generation process to produce $Q_{1A}$ and $Q_{2A}$ that contains simulated feature-level anomalies. $Q_{1A}$ and $Q_{2A}$ are then input into the subspace restriction module that attempts to restore the extracted features to an anomaly free representation $Q_{1S}$, $Q_{2S}$. Note that the anomaly generation process is performed only during training so at inference $Q_1$ and $Q_2$ are directly input into the subspace restriction module since they may already contain an anomaly during testing. 

For clarity, the steps that are only performed during training are marked with blue and steps only done at inference are marked with orange in Figure~\ref{fig:3dsr}, while the trainable modules are marked with red in Figure~\ref{fig:3dsr}. The Subspace restriction module is trained with an $L1$ loss to reconstruct anomaly-features $Q_1$ and $Q_2$ from $Q_{1A}$ and $Q_{2A}$, respectively. 

The Object specific decoder is trained to reconstruct the anomaly free appearance from the reconstructed features $Q_{1S}$ and $Q_{2S}$. The pretrained general appearance decoder reconstructs the anomaly appearance $I_G$ and $D_G$ from $Q_{1A}$ and $Q_{2A}$ at training or $Q_{1}$ and $Q_{2}$ at inference. Then, $I_G$,$D_G$,$I_S$ and $D_S$ are concatenated and input into the Anomaly detection module. The Anomaly detection module is trained to localize the simulated anomalies during training and real anomalies at test time. It directly outputs an anomaly segmentation mask $M_{out}$ and is trained using the Focal loss~\cite{lin2017focal}. Following \cite{zavrtanik2022dsr,zavrtanik2021draem}, the image-level anomaly score is estimated by first smoothing $M_{out}$ with a Gaussian filter and then taking the maximum value of the smoothed mask.

During training, anomalies are generated by modifying the quantized feature maps $Q_1$ and $Q_2$ as follows. First an anomaly map $M$ is generated by thresholding and binarizing a Perlin noise map, following previous works~\cite{zavrtanik2022dsr,zavrtanik2021draem}. $M$ is then resized to fit to the spatial dimensions of $Q_1$ and $Q_2$. Feature vectors of $Q_1$ and $Q_2$ in regions correspond to positive values of $M$ are replaced with feature vectors randomly sampled from codebooks $VQ_1$ and $VQ_2$ respectively, generating modified feature maps $Q_{1A}$ and $Q_{2A}$ that contain simulated anomalies.

\section{Experiments}
\textbf{Datasets.} 3DSR is evaluated on two recent 3D anomaly detection benchmarks, the MVTec3D dataset~\cite{mvtec3d} and the Eyecandies dataset~\cite{bonfiglioli2022eyecandies}, each containing 10 different object classes. The MVTec3D~\cite{mvtec3d} anomaly detection benchmark contains 4147 scans obtained by a high-resolution industrial 3D sensor that also acquires RGB data. Of the $4147$ scans, $894$ are anomalous containing various defects that are visible in either RGB or 3D data.  
3DSR is also evaluated on the Eyecandies dataset~\cite{bonfiglioli2022eyecandies}, a difficult rendered dataset with RGB and 3D anomalies. It contains $10000$ anomaly-free examples for training and $500$ examples for testing of which $250$ are anomalous.

\textbf{Evaluation problem setup.} The evaluation is divided into $3$ different problem setups, where only the depth (3D setup), RGB images (RGB setup) or both the depth and RGB images (3D+RGB) are used. Since 3DSR is a 3D and RGB+3D method, we provide DSR~\cite{zavrtanik2022dsr} results instead in the RGB setup.

\textbf{Evaluation metrics.} Anomaly localization and image-level anomaly detection capabilities of each method are evaluated. For the image-level detection the standard image-level AUROC metric is used. For anomaly localization the PRO metric~\cite{bergmann2019mvtec} is used. Additionally, the mean pixel-level AUROC is used for localization evaluation.

\subsection{Implementation details}
In the first stage of training, the DADA module is trained using the simulated depth data for 3D and the ImageNet dataset~\cite{deng2009imagenet} for RGB supervision. DADA is trained using a batch size of $64$ for $100K$ iterations using a learning rate of $0.0002$. The vector quantization codebooks contains $2048$ embeddings of dimension $256$. In the second stage, 3DSR is trained on the MVTec3D dataset~\cite{mvtec3d} or the Eyecandies dataset~\cite{bonfiglioli2022eyecandies}. In both datasets the 3D data is given in the form of a sorted point cloud. The data is preprocessed by first normalizing the depth map to values between $0$ and $1$. Then, the missing values in each depth image are replaced with the average of all the valid pixels in a $3\times3$ neighborhood. If there are no surrounding valid pixels, the value is set to 0. A foreground mask is obtained by classifying points as either foreground or background by it's distance to the background plane. The background plane equation is obtained from valid points at the edges of the sorted point cloud. During training, anomalies are generated only on the foreground object. 3DSR is trained on an individual object class of each dataset as is standard for surface anomaly detection methods~\cite{zavrtanik2021draem,zavrtanik2022dsr,roth2021towards,rudolph2023ast}. It is trained with a batch size of $16$ for $30K$ iterations with a learning rate of $0.0002$.

\subsection{Results}
\begin{table*}[htbp]
    \centering
    \caption{Anomaly detection results on the MVTec3D dataset for the 3D, RGB and 3D+RGB problem setups. The results are listed as image-level AUROC scores (higher is better). The results of evaluated methods are ranked and the first, second and third place are marked.}
    \resizebox{0.9 \linewidth}{!}{\begin{tabular}{cccccccccccc|c}
        \toprule
        & Method & Bagel & Cable Gland & Carrot & Cookie & Dowel & Foam & Peach & Potato & Rope & Tire & Mean\\
        \hline
        \multirow{10}{*}{\rotatebox[origin=c]{90}{3D}} & Voxel AE \cite{mvtec3d} & 69.3 & 42.5 & 51.5 & 79.0 & 49.4 & 55.8 & 53.7 & 48.4 & 63.9 & 58.3 & 57.1 \\
        & Depth GAN \cite{mvtec3d} & 53.0 & 37.6 & 60.7 & 60.3 & 49.7 & 48.4 & 59.5 & 48.9 & 53.6 & 52.1 & 52.3 \\
        & Depth AE \cite{mvtec3d} & 46.8 & \second{73.1} & 49.7 & 67.3 & 53.4 & 41.7 & 48.5 & 54.9 & 56.4 & 54.6 & 54.6 \\
        & FPFH~\cite{horwitz2022empirical} & 82.5 & 55.1 & 95.2 & 79.7 & \third{88.3} & 58.2 & 75.8 & 88.9 & 92.9 & \third{65.3} & 78.2 \\
        & 3D-ST~\cite{bergmann2023ST} & 86.2 & 48.4 & 83.2 & \third{89.4} & 84.8 & 66.3 & 76.3 & 68.7 & \second{95.8} & 48.6 & 74.8 \\
        & AST$_{3D}$~\cite{rudolph2023ast} & \third{88.1} & 57.6 & \second{96.5} & \second{95.7} & 67.9 & \second{79.7} & \first{99.0} & \third{91.5} & \third{95.6} & 61.1 & \third{83.3} \\
        & M3DM$_{3D}$~\cite{wang2023m3dm} & \second{94.1} & \third{65.1} & \second{96.5} & \first{96.9} & \second{90.5} & \third{76.0} & \third{88.0} & \first{97.4} & 92.6 & \second{76.5} & \second{87.4} \\
        & 3DSR$_{3D}$ & \first{94.5} & \first{83.5} & \first{96.9} & 85.7 & \first{95.5} & \first{88.0} & \second{96.3} & \third{93.4} & \first{99.8} & \first{88.8 }& \first{92.2} \\
        \hline
        \multirow{10}{*}{\rotatebox[origin=c]{90}{RGB}} &PatchCore~\cite{roth2021towards} & 87.6 & 88.0 & 79.1 & 68.2 & 91.2 & 70.1 & 69.5 & 61.8 & 84.1 & 70.2 & 77.0 \\
        & DifferNet~\cite{rudolph2021same} & 85.9 & 70.3 & 64.3 & 43.5 & 79.7 & 79.0 & 78.7 & \third{64.3} & 71.5 & 59.0 & 69.6 \\
        & PADiM~\cite{defard2021padim} & \first{97.5} & 77.5 & 69.8 & 58.2 & 95.9 & 66.3 & 85.8 & 53.5 & 83.2 & 76.0 & 76.4 \\
        & CS-Flow~\cite{rudolph2022csflow} & 94.1 & \first{93.0} & 82.7 & \second{79.5} & \second{99.0} & \third{88.6} & 73.1 & 47.1 & \second{98.6} & 74.5 & 83.0 \\
        & AST$_{RGB}$~\cite{rudolph2023ast} & \second{94.7} & \third{92.8} & \third{85.1} & \first{82.5} & \third{98.1} & \first{95.1} & \third{89.5} & 61.3 & \first{99.2} & \second{82.1} & \second{88.0} \\
        & M3DM$_{RGB}$~\cite{wang2023m3dm} & \third{94.4} & 91.8 & \second{89.6} & 74.9 & 95.9 & 76.7 & \second{91.9} & \second{64.8} & 93.8 & \third{76.7} & \third{85.0} \\
        & DSR$_{RGB}$~\cite{zavrtanik2022dsr} & 84.4 & \first{93.0} & \first{96.4} & \third{79.4} & \first{99.8} & \second{90.4} & \first{93.8} & \first{73.0} & \third{97.8} & \first{90.0} & \first{89.8} \\
        \hline
        \multirow{8}{*}{\rotatebox[origin=c]{90}{3D+RGB}} & Voxel AE \cite{mvtec3d} & 51.0 & 54.0 & 38.4 & 69.3 & 44.6 & 63.2 & 55.0 & 49.4 & 72.1 & 41.3 & 53.8 \\ 
        & Depth GAN \cite{mvtec3d} & 53.8 & 37.2 & 58.0 & 60.3 & 43.0 & 53.4 & 64.2 & 60.1 & 44.3 & 57.7 & 53.2 \\
        & Depth AE \cite{mvtec3d} & 64.8 & 50.2 & 65.0 & 48.8 & 80.5 & 52.2 & 71.2 & 52.9 & 54.0 & 55.2 & 59.5 \\
        & PatchCore+FPFH~\cite{horwitz2022empirical} & 91.8 & 74.8 & 96.7 & 88.3 & 93.2 & 58.2 & 89.6 & \third{91.2} & 92.1 & \third{88.6} & 86.5 \\
        & AST~\cite{rudolph2023ast} & \second{98.3} & \second{87.3} & \second{97.6} & \third{97.1} & \third{93.2} & \third{88.5} & \second{97.4} & \first{98.1} & \first{100} & 79.7 & \third{93.7} \\
        & M3DM~\cite{wang2023m3dm} & \first{99.4} & \first{90.9} & \third{97.2} & \second{97.6} & \second{96.0} & \second{94.2} & \third{97.3} & 89.9 & \third{97.2} & \third{85.0} & \second{94.5} \\
        & 3DSR & \third{98.1} & \third{86.7} & \first{99.6} & \first{98.1} & \first{100} & \first{99.4} & \first{98.6} & \second{97.8} & \first{100} & \first{99.5} & \first{97.8} \\
        \bottomrule
    \end{tabular}}
    \label{tab:auroc_img}
\end{table*}

\begin{table*}[htbp]
    \centering
    \caption{Anomaly localization results on the MVTec3D dataset for the 3D, RGB and 3D+RGB problem setups. The results are listed as AUPRO scores (higher is better). The results of evaluated methods are ranked and the first, second and third place are marked.}
    \resizebox{0.9 \linewidth}{!}{\begin{tabular}{cccccccccccc|c}
        \toprule
        & Method & Bagel & Cable Gland & Carrot & Cookie & Dowel & Foam & Peach & Potato & Rope & Tire & Mean\\
        \hline
        \multirow{5}{*}{\rotatebox[origin=c]{90}{3D}} & Depth AE \cite{mvtec3d} & 14.7 & 6.9 & 29.3 & 21.7 & 20.7 & 18.1 & 16.4 & 6.6 & 54.5 & 14.2 & 20.3 \\
        &Depth GAN \cite{mvtec3d} & 11.1 & 7.2 & 21.2 & 17.4 & 16.0 & 12.8 & 0.3 & 4.2 & 44.6 & 7.5 & 14.3 \\
        & Voxel AE \cite{mvtec3d} & 26.0 & 34.1 & 58.1 & 35.1 & 50.2 & 23.4 & 35.1 & 65.8 & 1.5 & 18.5 & 34.8 \\
        & FPFH \cite{horwitz2022empirical} & \first{97.3} & \first{87.9} & \second{98.2} & \first{90.6} & \second{89.2} & \second{73.5} & \first{97.7} & \first{98.2} & \second{95.6} & \first{96.1} & \first{92.4} \\
        & M3DM~\cite{wang2023m3dm} & \second{94.3} & \third{81.8} & \third{97.7} & \second{88.2} & \third{88.1} & \first{74.3} & \third{95.8} & \third{97.4} & \third{95.0} & \second{92.9} & \third{90.6} \\
        & 3DSR & \third{92.2} & \second{87.2} & \first{98.4} & \third{85.9} & \first{94.0} & \third{71.4} & \second{97.0} & \second{97.8} & \first{97.7} & \third{85.8} & \second{90.7} \\
        \hline
        \multirow{5}{*}{\rotatebox[origin=c]{90}{RGB}} &CFlow \cite{gudovskiy2022cflow} & 85.5 & 91.9 & \third{95.8} & 86.7 & 96.9 & 50.0 & 88.9 & 93.5 & 90.4 & \third{91.9} & 87.1 \\
        & PatchCore \cite{roth2021towards} & 90.1 & \third{94.9} & 92.8 & 87.7 & 89.2 & 56.3 & 90.4 & 93.2 & 90.8 & 90.6 & 87.6 \\
        & PADiM \cite{mvtec3d} & \first{98.0} & 94.4 & 94.5 & \first{92.5} & \third{96.1} & \third{79.2} & \second{96.6} & \third{94.0} & \third{93.7} & 91.2 & \third{93.0} \\
        & M3DM~\cite{wang2023m3dm} & \second{95.2} & \first{97.2} & \second{97.3} & \second{89.1} & 93.2 & \second{84.3} & \first{97.0} & \first{95.6} & \first{96.8} & \second{96.6} & \second{94.2} \\
        & DSR~\cite{zavrtanik2022dsr} & \third{92.3} & \second{97.0} & \first{97.9} & 85.9 & \first{97.9} & \first{89.4} & \third{94.3} & \second{95.1} & \second{96.4} & \first{98.0} & \first{94.4} \\
        \hline
        \multirow{8}{*}{\rotatebox[origin=c]{90}{3D+RGB}} & Depth AE \cite{mvtec3d} & 43.2 & 15.8 & 80.8 & 49.1 & 84.1 & 40.6 & 26.2 & 21.6 & 71.6 & 47.8 & 48.1 \\
        & Depth VM \cite{mvtec3d} & 38.8 & 32.1 & 19.4 & 57.0 & 40.8 & 28.2 & 24.4 & 34.9 & 26.8 & 33.1 & 33.5 \\
        & Voxel AE \cite{mvtec3d} & 46.7 & 75.0 & 80.8 & 55.0 & 76.5 & 47.3 & 72.1 & 91.8 & 1.9 & 17.0 & 56.4 \\
        & 3D-ST~\cite{bergmann2023ST} & 95.0 & 48.3 & \first{98.6} & 92.1 & 90.5 & 63.2 & 94.5 & \first{98.8} & \second{97.6} & 54.2 & 83.3 \\
        & PatchCore + FPFH \cite{horwitz2022empirical} & \first{97.6} & \second{96.9} & \third{97.9} & \first{97.3} & \third{93.3} & \third{88.8} & \third{97.5} & \second{98.1} & 95.0 & \third{97.1} & \third{95.9} \\
        & M3DM~\cite{wang2023m3dm} & \second{97.0} & \first{97.1} & \third{97.9} & \second{95.0} & \second{94.1} & \second{93.2} & \second{97.7} & 97.1 & \third{97.1} & \second{97.5} & \second{96.4} \\
        & 3DSR & \third{96.4} & \third{96.6} & \second{98.1} & \third{94.2} & \first{98.0} & \first{97.3} & \first{98.1} & \third{97.7} & \first{97.9} & \first{97.9} & \first{97.2} \\
        \bottomrule
    \end{tabular}}
    \label{tab:aupro}
\end{table*}

Anomaly detection results in terms of image-level AUROC are shown in Table~\ref{tab:auroc_img}. 3DSR significantly outperforms competing methods in the 3D settings, where only depth information is used. The 3D image-level AUROC is improved by approximately $5$ percentage points. This demonstrates the informative depth representations learned by DADA and validates the 3DSR pipeline.

In classes where depth information is vital for anomaly detection, such as Tire and Foam, the improvement in the 3D setting is even greater. Additionally, 3DSR outperforms competing methods in the 3D+RGB setup by $3.3$ percentage points demonstrating the ability of DSR to efficiently utilize information from both the depth and RGB modalities again emphasizing the powerful joint representations of 3D and RGB learned by DADA using the proposed data simulation process. 
In Table~\ref{tab:auroc_img} the first, second and third best performing methods are marked for each object class. Note that 3DSR achieves first place on $7$ out of $10$ classes for the 3D setup and $5$ out of $10$ classes on the 3D+RGB setup, while staying in the top 3 in every object class. 

Anomaly localization results are shown in Table \ref{tab:aupro} in terms of the AUPRO metric~\cite{bergmann2019mvtec}. 3DSR achieves second place in segmentation results. It achieves comparable results to M3DM in the 3D problem setup while outperforming it in the RGB+3D setup.
A comparison of 3DSR and state-of-the-art methods in terms of mean pixel-level and image-level AUROC on the RGB+3D setup is shown in Table~\ref{tab:pixauc}, where 3DSR outperforms both AST~\cite{rudolph2023ast} and M3DM~\cite{wang2023m3dm} on both image-level and pixel-level AUROC.

\begin{table}[htbp]
    \centering
    \resizebox{0.7 \linewidth}{!}{
    \begin{tabular}{ccc}
        \toprule
        Method & I-AUROC & P-AUROC \\
        \midrule
        PatchCore + FPFH~\cite{horwitz2022empirical} & 86.5 & 99.2 \\
        AST~\cite{rudolph2023ast} & 93.7 & 97.6 \\
        M3DM~\cite{wang2023m3dm} & 94.5 & 99.2 \\
        3DSR & \textbf{97.8} & \textbf{99.5} \\
        \bottomrule
    \end{tabular}}
    \caption{Mean image-level AUC and pixel-level AUC values on the 3D+RGB setup on the MVTec3D dataset.}
    \label{tab:pixauc}
\end{table}

\begin{table*}[htbp]
    \centering
    \resizebox{1.0 \linewidth}{!}{\begin{tabular}{ccccccccccc|c}
\toprule
Method & Candy & Chocolate & Chocolate & Confetto & Gummy & Hazelnut & Licorice & Lollipop & Marshmallow & Peppermint & Mean\\
& cane & cookie & praline & & Bear & truffle & sandwich &  &  & candy & \\
\hline
3DSR$_{3D}$ & 60.0 & 76.8 & 74.2 & 77.0 & 76.1 & 74.9 & 81.1 & 83.1 & 81.1 & 91.7 & \textbf{77.6} \\
M3DM$_{3D}$ & 48.2 & 58.9 & 80.5 & 84.5 & 78.0 & 53.8 & 76.6 & 82.7 & 80.0 & 82.2 & 72.5 \\

\hline
DSR$_{RGB}$~\cite{zavrtanik2022dsr} & 70.6 & 96.5 & 95.0 & 96.6 & 87.0 & 79.0 & 88.5 & 85.7 & 99.8 & 99.2 & \textbf{89.8} \\
M3DM$_{RGB}$ & 64.8 & 94.9 & 94.1 & 100 & 87.8 & 63.2 & 93.3 & 81.1 & 998 & 100 & 87.9 \\

\hline
3DSR$_{3D+RGB}$ & 65.1 & 99.8 & 90.4 & 97.8 & 87.5 & 86.1 & 96.5 & 89.9 & 99.0 & 97.1 & \textbf{90.9} \\
M3DM$_{3D+RGB}$ & 62.4 & 95.8 & 95.8 & 100 & 88.6 & 75.8 & 94.9 & 83.6 & 100 & 100 & 89.7 \\

\bottomrule
\end{tabular}}
    \caption{Comparison between M3DM~\cite{wang2023m3dm} and 3DSR on the Eyecandies dataset in terms of image-level AUROC (higher is better).}
    \label{tab:eyecan}
\end{table*}

Table~\ref{tab:eyecan} shows the comparison between 3DSR and the previous top performing method M3DM~\cite{wang2023m3dm} on the Eyecandies~\cite{bonfiglioli2022eyecandies} dataset in terms of the image-level AUROC on the 3D and 3D+RGB problem setups. 3DSR outperforms M3DM~\cite{wang2023m3dm} on the 3D anomaly detection setup and achieves an image-level AUROC score that is $3$ percentage points higher than that of M3DM~\cite{wang2023m3dm}. On the 3D+RGB anomaly detection setup 3DSR achieves state-of-the-art performance, slightly outperforming M3DM~\cite{wang2023m3dm} in the mean AUROC score, while achieving a lower variance in scores across classes. The results suggest that in the Eyecandies dataset~\cite{bonfiglioli2022eyecandies} most anomalies that are perceptible in 3D are also visible in RGB, since similar results are achieved in the RGB and the 3D+RGB problem setups.

\begin{figure*}[htbp]
  \centering
  \includegraphics[width=0.95\linewidth]{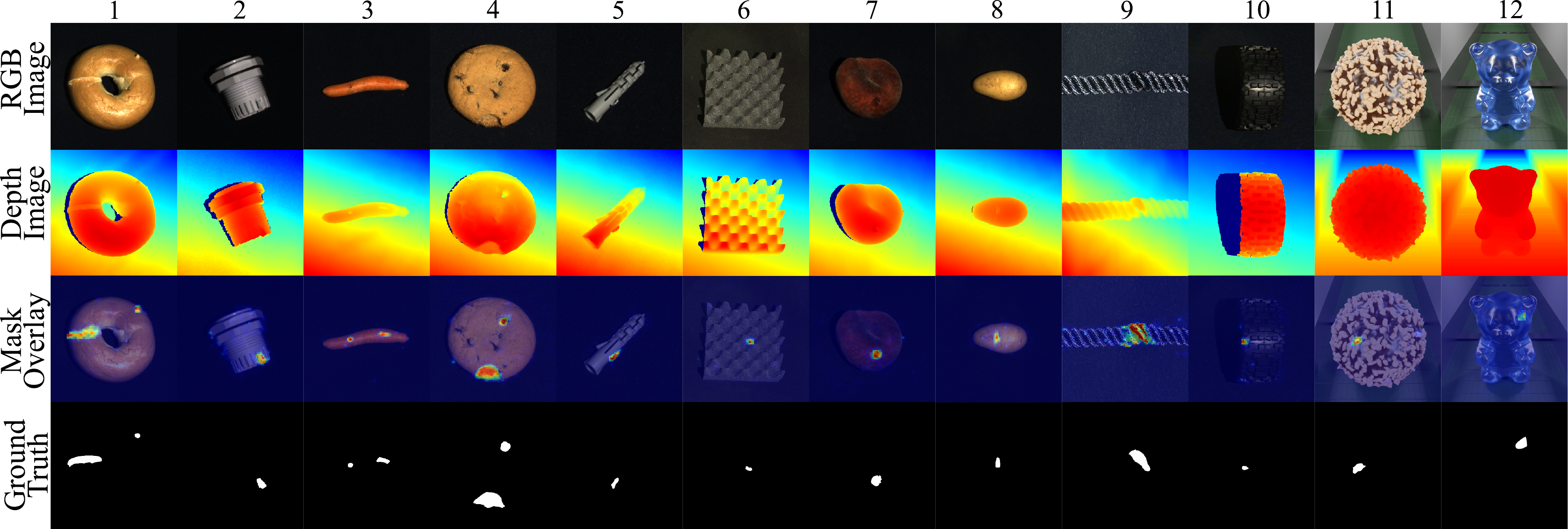}
  \caption{Qualitative results of 3DSR on the MVTec3D and Eyecandies benchmarks.}
  \label{fig:examples}
\end{figure*}

In Figure~\ref{fig:examples}, qualitative examples for the MVTec3D~\cite{mvtec3d} and Eyecandies~\cite{bonfiglioli2022eyecandies} datasets are shown. The first two rows contain the RGB and depth images, respectively. The third row shows the 3DSR output mask overlaid on the RGB image. The ground truth anomaly masks are shown in row 4. In classes such as Cookie (Column 4), Peach (Column 7) and Potato (Column 8), anomalies are subtle in the RGB image, but are visible in the depth image and are detected by 3DSR. In Column 6, the anomaly on the foam is only visible in the RGB space and is also detected by 3DSR. In other columns the anomalies are visible in both RGB and depth images. The last two columns show examples from the Eyecandies dataset~\cite{bonfiglioli2022eyecandies}. Column 11 contains a 3D anomaly and column 12 contains an anomaly only visible in the RGB image. 3DSR segments all of these examples accurately.

\section{Ablation study}
The results of the ablation study are shown in Table~\ref{tab:ab}. First, the naive approach of training the VQVAE of DSR~\cite{zavrtanik2022dsr} on RGB+depth image pairs of all classes in the MVTec3D~\cite{mvtec3d} training set is evaluated in experiment $DSR_{naive}$. This results in poor anomaly detection performance showing the need for better RGB+3D representations enabled by the proposed contributions. 

The contribution of the simulated training data is evaluated in the $3DSR_{no\_affine}$ and $3DSR_{no\_perlin}$ experiments. In the $3DSR_{no\_perlin}$ experiment the use of Perlin noise maps for training DADA is omitted and the depth training data is replaced with ImageNet images converted to grayscale. Grayscale images do not sufficiently model the properties of depth images and the image-level AUROC score drops significantly by approximately $8$ percentage points. In $3DSR_{no\_affine}$ only the perlin noise map scaled from $0$ to $1$ is used without the additional scaling with $\alpha$ and $\beta$ as described in Section~\ref{sec:depthgen}. This leads to an approximately $3$ percentage point drop in the image-level AUROC. Experiments $3DSR_{no\_affine}$ and $3DSR_{no\_perlin}$ demonstrate the effectiveness of using Perlin noise as a simulation source of industrial depth and the benefit of using an affine transformation of the simulated data to model the characteristics of the data. 

The contribution of the DADA module is evaluated in experiments $3DSR_{VQVAE}$, $3DSR_{weighted}$. In these experiments, a VQVAE from \cite{zavrtanik2022dsr} is used to learn the joint RGB and depth representations, where the RGB and depth representations are not separated by grouped convolutions. In $3DSR_{VQVAE}$ the proposed DADA module is replaced with a VQVAE~\cite{vqvae2} model which causes an approximately $2$ percentage point drop in image-level AUROC. This experiment shows the benefit of separating the RGB and depth data in the DADA architecture and shows that this separation leads to an improved downstream anomaly detection. Experiment $3DSR_{weighted}$ also uses a vector quantized autoencoder from \cite{zavrtanik2022dsr} but the $\lambda_{D}$ and $\lambda_{I}$ values in the loss in Equation (\ref{eq:vqv}) are set to the best-performing values $\lambda_{D}=10$ and $\lambda_{I}=1$, increasing the loss contribution of depth image reconstruction. This leads to a $1.3$ percentage point drop in image-level AUROC suggesting that replacing DADA with VQ-VAE and a simple loss reweighing is not sufficient. The change in $\lambda_{I}$,$\lambda_{D}$ also accounts for the difference between $3DSR_{weighted}$ and $3DSR_{VQVAE}$, showing the impact of the choice of $\lambda_{I}$,$\lambda_{D}$.
\begin{table}[htbp]
    \centering
    \resizebox{0.8 \linewidth}{!}{
    \begin{tabular}{cccc}
        \toprule
        Method & I-AUROC & P-AUROC & PRO\\
        \midrule
        DSR$_{naive}$ & 87.6 & 96.5 & 92.3 \\
        3DSR$_{no\_perlin}$ & 90.0 & 98.3 & 93.3\\
        3DSR$_{no\_affine}$ & 94.8 & 99.2 & 95.9 \\
        3DSR$_{VQVAE}$ & 95.8 & 99.3 & 96.3 \\
        3DSR$_{weighted}$ & 96.5 & 99.4 & 96.7 \\
        3DSR & \textbf{97.8} & \textbf{99.5} & \textbf{97.2} \\
        \bottomrule
    \end{tabular}}
    \caption{Ablation study results.}
    \label{tab:ab}
\end{table}

\textbf{Inference efficiency.} The performance in terms of frames-per-second (FPS) is evaluated on an NVIDIA RTX A4500 GPU and shown in Table~\ref{tab:fps}. Compared to other recent methods such as M3DM~\cite{wang2023m3dm}, 3DSR is very fast and can run in real-time on a GPU due to the efficiency of the DADA and the anomaly segmentation module. The previous best method M3DM~\cite{wang2023m3dm} that uses point-cloud data requires heavy preprocessing and two large transformer networks. 3DSR is an order of magnitude faster than M3DM and is also almost twice as fast as AST~\cite{rudolph2023ast} in terms of FPS.
\begin{table}[htbp]
    \centering
    \resizebox{0.7 \linewidth}{!}{
    \begin{tabular}{cccc}
        \toprule
        Method & AST~\cite{rudolph2023ast} & M3DM~\cite{wang2023m3dm} & 3DSR \\
        \midrule
        FPS & \second{18} & \third{0.6} & \first{33} \\
        \bottomrule
    \end{tabular}}
    \caption{Method performance in terms of frames-per-second (FPS) on the NVIDIA RTX A4500 GPU.}
    \label{tab:fps}
\end{table}

\begin{figure}[htbp]
  \centering
  \includegraphics[width=0.75\linewidth]{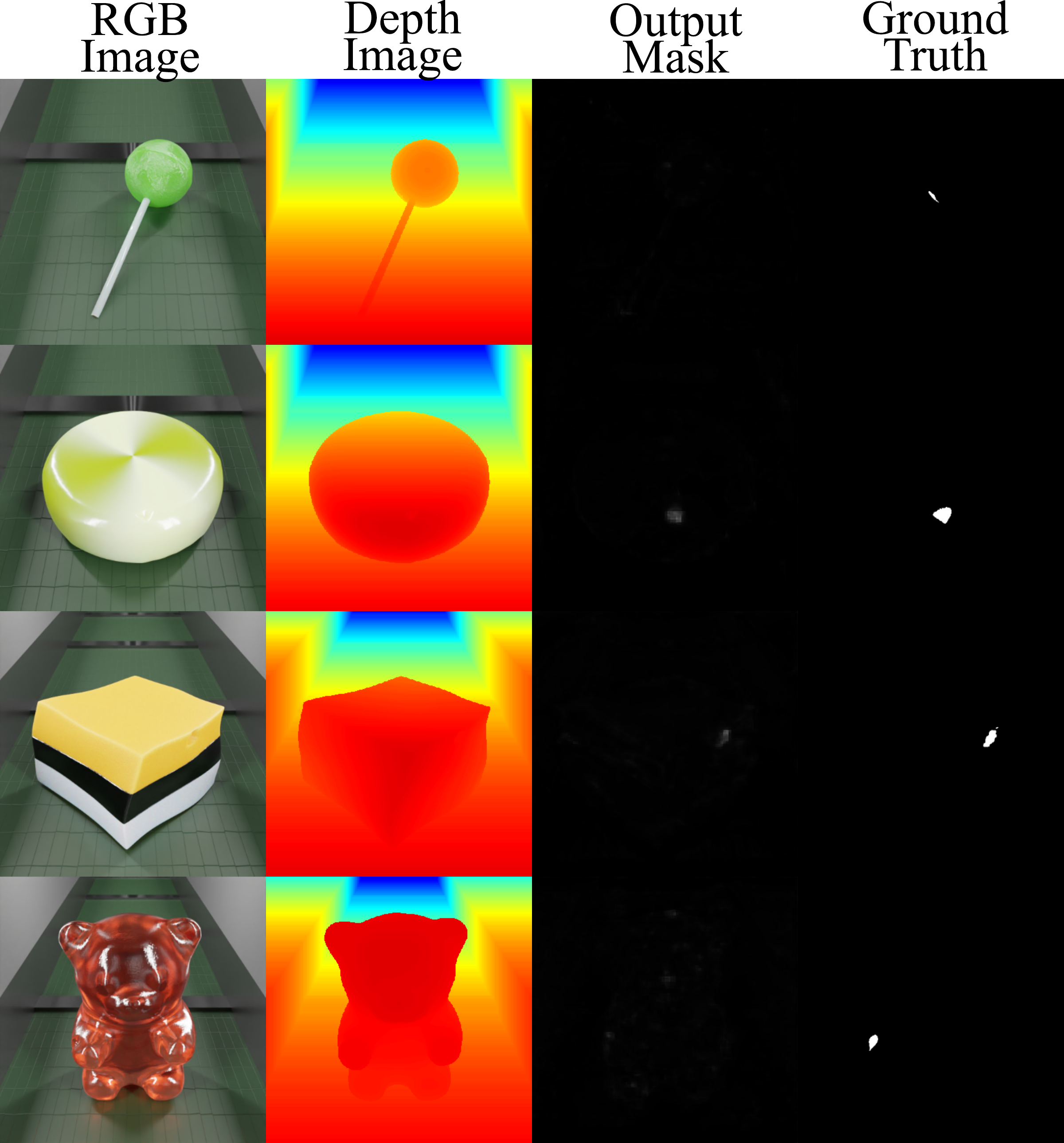}
  \caption{Difficult cases on the Eyecandies dataset.}
  \label{fig:fail_cases}
\end{figure}

\textbf{Limitations.} Figure~\ref{fig:fail_cases} shows Eyecandies dataset~\cite{bonfiglioli2022eyecandies} examples that are particularly difficult for 3DSR. In row one, the anomaly is a small dent at the object edge. It is not visible in the RGB image or distinguishable from the natural edge in the depth map making it difficult to detect. In rows 2 and 3, the anomalies are depth-based and result in minor changes to the object's surface. They are hardly visible in the RGB images. Nonetheles, 3DSR is able to segment them with a lower confidence. In row 4, the anomaly is a deformation visible in the RGB image, however due to the object's transparency, such anomalies are difficult to detect. Note that the Eyecandies dataset~\cite{bonfiglioli2022eyecandies} contains challenging anomalies that closely resemble the object's normal appearance, making them difficult to detect, even for humans.

\section{Conclusion}

We proposed a 3D anomaly detection method 3DSR, which is capable of detecting 3D anomalies in industrial depth data and can even utilize depth and RGB data to further improve the anomaly detection performance. Our first contribution is the novel Depth-aware Discrete Autoencoder (DADA) that separately encodes 3D and RGB data during training thus learning better representations of individual modalities. The second contribution is the simulated depth generation process for learning robust representations of industrial 3D data. The new method 3DSR (third contribution), achieves state-of-the-art results on the MVTec3D~\cite{mvtec3d} and Eyecandies~\cite{bonfiglioli2022eyecandies} datasets. On the MVTec3D anomaly detection dataset~\cite{mvtec3d}, 3DSR surpasses competing methods significantly in the 3D and 3D+RGB anomaly detection setups demonstrating a strong 3D anomaly detection capability validating the proposed contributions. 3DSR is faster than competing methods and is an order of magnitude faster than the previous best method M3DM~\cite{wang2023m3dm} which uses a point-cloud-based 3D information extraction. The proposed depth simulation may also help transfer the recent progress in RGB anomaly detection methods to the 3D domain by improving the representations extracted by backbone networks by training on simulated data with self-supervision. %We leave this for future work.

\noindent \textbf{Acknowledgement} \textit{This work was supported by Slovenian research agency programs J2-3169, J2-2506, P2-0214 and 23-20MR.R588}
%-------------------------------------------------------------------------
{\small
\bibliographystyle{ieee_fullname}
\bibliography{egbib}
}

\end{document}